\documentclass[dvipsnames,format=sigconf,anonymous=false,review=false]{acmart}
\AtBeginDocument{%
  }

\setcopyright{acmlicensed}
\copyrightyear{2026}
\acmYear{2026}
\setcopyright{acmlicensed}\acmConference[GECCO '26]{Genetic and
Evolutionary Computation Conference}{July 13-17, 2026}{San Jos\'e, Costa Rica}
\acmBooktitle{Genetic and Evolutionary Computation Conference (GECCO '26),
July 13-17, 2026}
\acmDOI{XXXXXXX}
\acmISBN{XXXXXX}

\usepackage{algorithm}
\usepackage{algpseudocode}
\usepackage{xcolor}
\usepackage{float}
\usepackage{booktabs}
\usepackage{siunitx}
\usepackage{subcaption}
\usepackage{multirow}
\usepackage{comment}
\usepackage{threeparttable}
\usepackage{caption}
\usepackage{eso-pic}

\AddToShipoutPictureFG*{%
\AtPageUpperLeft{%
\raisebox{-1.5cm}[0pt][0pt]{%
\makebox[\paperwidth]{%
\hfill
\textcolor{red}{This is the first submitted version of the manuscript.}
\hfill
}}}}
\begin{document}

\title{Energy-Aware Metaheuristics}

\author{Enrique Alba}
\affiliation{%
  \institution{University of M\'{a}laga}
  \city{M\'{a}laga}
  \country{Spain}}
\email{ealbat@uma.es}

\author{Tomohiro Harada}
\affiliation{%
  \institution{Saitama University}
  \city{Saitama}
  \country{Japan}}
\email{tharada@mail.saitama-u.ac.jp}

\author{Gabriel Luque}
\affiliation{%
 \institution{University of M\'{a}laga}
 \city{M\'{a}laga}
 \country{Spain}}
\email{gluque@uma.es}

\begin{abstract}
This paper presents a principled framework for designing energy-aware metaheuristics that operate under fixed energy budgets. We introduce a unified operator-level model that quantifies both numerical gain and energy usage, and define a robust Expected Improvement per Joule (EI/J) score that guides adaptive selection among operator variants during the search. The resulting energy-aware solvers dynamically choose between operators to self-control exploration and exploitation, aiming to maximize fitness gain under limited energy. We instantiate this framework in three representative metaheuristics---steady-state GA, PSO, and ILS---each equipped with lightweight and heavy operator variants. 
Experiments on three heterogeneous combinatorial problems (Knapsack, NK-landscapes, and Error-Correcting Codes) show that the energy-aware variants consistently reach comparable fitness while requiring substantially less energy than their non-energy-aware baselines. EI/J values stabilize early and yield clear operator-selection patterns, with each solver reliably self-identifying the most improvement-per-Joule–efficient operator across problems. 
\end{abstract}




\maketitle

\section{Introduction}

Energy consumption is emerging as a critical bottleneck in modern software and AI. Solving complex problems with optimisation and learning techniques usually leads to intensive computing, and even if different approaches are explored~\cite{BOLONCANEDO2024128096,Tabbakh2024,Muralidhar2022}, the algorithms driving the search for a solution receive indirect (though important) approaches to reduce their numerical steps~\cite{Li2022PHEV,Harada2024Energy}, time/memory~\cite{Andoorveedu2022,Brain2023}, and implementation refinements~\cite{Abdelhafez2019,Escobar2019}.

Arguably, ignoring green computing during the design of a technique needs to change, to become the ``classic" way of building algorithms, meaning that we need to include energy-aware techniques as an integral part of any new technique. This should be a standard approach, admitting variants such as completely disabling the energy part to proceed with purely numerical or physical studies, just like until now. We face a future with energy considerations in algorithmic operations, enabling a scientifically scalable and socially acceptable contribution to research.

Metaheuristics are widely used in AI for their robustness and domain-agnostic nature, but they traditionally assume unlimited energy. As computing shifts toward embedded, battery-powered, and large-scale parallel systems with explicit energy limits, designing \emph{energy-aware} metaheuristics becomes necessary.

Current approaches to energy-aware optimisation mostly focus on hardware-level techniques \cite{Mereno2016,Meier2025} or large-scale algorithmic simplifications~\cite{Menghani2023,Liu2024LightDL,Tmamna2024}. However, metaheuristics offer a unique opportunity: their behavior relies on modular operators (mutation, crossover, velocity updates, neighborhood search, acceptance) whose computational cost varies widely and can be profiled independently. This suggests that energy awareness can be introduced \emph{at the operator level}, enabling fine-grained numerical control for users. We here propose to answer three main research questions:
\begin{enumerate}
    \item Q1: Can we offer profiling data to start comparisons on energy in papers beyond numerical performance?
    \item Q2: How can a grounded principled mechanism for energy-aware search of solutions be built, combining energy awareness with expected numerical improvement? 
    \item Q3: What patterns do emerge across algorithms and problems from combining numerical and energy considerations?
\end{enumerate}

Q1 finds experimental data that later becomes information (stats) and then knowledge (mathematical modelling). Q2 makes sense, since most algorithms treat operators as energy-free interchangeable components. Q3 naturally poses the challenge of understanding and using new AI solvers across problems. To address these challenges, we propose a framework for energy-aware metaheuristics based on three ideas linked to our study cases:

\begin{itemize}
    \item \textbf{Algorithm-level energy profiling:} For each of the used GA, PSO, and ILS versions (lighter and heavier), we measure energy usage to offer insights for future statistical models and comparisons.
    \item \textbf{Energy-based operator efficiency:} We define \emph{Expected Improvement per Joule} (EI/J), a ratio between the expected numerical improvement an operator produces and its expected energy expenditure.
    \item \textbf{Higher level understanding:} compare algorithms and problems to get a wide point of view on the internals and externals of different techniques.
\end{itemize}

In short, we construct energy-aware versions of GA, PSO, and ILS by incorporating the scheduler into the main loop and replacing uniform operator invocation with energy-adaptive decisions. Empirical results show that these algorithms retain the search properties and solution quality of their classic counterparts, while offering explicit control over energy consumption.

The remainder of the paper describes the methodology, the theoretical model of expected improvement per Joule, the energy profiling of operators, and the experimental results.


\section{Background and Related Work}




Energy measurement in software systems has been approached using hardware- and software-based techniques~\cite{noureddine-ause-2015}. External power meters provide accurate measurements but are costly, intrusive, and hard to scale~\cite{Muralidhar2022}. For commodity CPUs, on-chip sensors such as Intel’s Running Average Power Limit (RAPL) enable low-overhead access to CPU/DRAM energy and are widely used in empirical studies~\cite{David2010,pyrapl_docs,Abdelhafez2019}. Software-level tools based on estimation models and carbon-intensity proxies (e.g., CodeCarbon) focus on coarse-grained energy/CO$_2$ reporting and lack the temporal precision required for operator-level analysis~\cite{codecarbon_zenodo,BOLONCANEDO2024128096}.

Energy consumption in optimisation and metaheuristics has been analysed across evolutionary and swarm-based methods~\cite{Abdelhafez2019,ABDELHAFEZ2020100692,Harada2024Energy}. These works typically report energy alongside runtime and solution quality~\cite{Jamil2022,MeleloGuervos2024,Cotta2024}, showing that algorithmic and parameter choices (e.g., population size) can significantly affect energy behaviour even when solution quality is similar~\cite{Diaz2022,Fernandez2020}. However, energy is generally treated as an external metric and does not guide the search process~\cite{Jamil2022,Verdecchia2023}.

Metaheuristics are naturally structured as compositions of operators, motivating adaptive control, operator selection, and hyper-heuristic approaches~\cite{Fialho2010AOS,Burke2013}. These mechanisms estimate operator utility online using numerical improvement and performance feedback~\cite{Fialho2010AOS,Burke2013}. Yet, they typically assume comparable computational or energy costs across operators, an assumption that breaks down when operators differ substantially in complexity or constant factors~\cite{Abdelhafez2019,Jamil2022}.

Energy-constrained optimisation can be framed as resource-bounded search, where the budget is expressed in energy rather than time or iterations~\cite{Zilberstein1996,Verdecchia2023}. While anytime and \sloppy budgeted-computation studies address trade-offs between computation and solution quality, they rarely rely on energy measurements with sufficient granularity for operator-level control~\cite{Challa2026AnytimeSurvey,Muralidhar2022}. As predictive models of numerical improvement are difficult to obtain a priori, adaptive approaches often learn expected improvement empirically during execution~\cite{Fialho2010AOS,Burke2013}, motivating selection criteria that relate numerical gain to measured energy expenditure~\cite{Abdelhafez2019,Harada2024Energy}.

Overall, existing work focuses mainly on measuring or analysing energy consumption. Using energy information to drive online algorithmic decisions—especially at the operator level—remains largely unexplored and is the focus of this study.











\section{Expected Improvement per Joule (EI/J)}
\label{sec:eij}

This section introduces the \emph{Expected Improvement per Joule} (EI/J) criterion,
a quantitative measure of the expected numerical benefit per unit of energy spent
when applying a metaheuristic operator. EI/J provides the foundation for our
energy-aware operator scheduler and allows metaheuristics to dynamically
prioritise operators that deliver the highest efficiency under a fixed or
progressively shrinking energy budget. The concept is applicable to any
metaheuristic and any operator whose improvement and energy cost can be
measured or estimated.

\subsection{Motivation}

Classical metaheuristics assume that all operators have equal cost: mutation and crossover in GA, velocity updates in PSO, and neighborhood explorations in ILS are usually treated as computationally equivalent actions. In reality, their energy
consumption differs by orders of magnitude depending on their algorithmic complexity and low-level software/hardware behaviour. Similarly, operators differ in their expected fitness improvement, which may depend on the search state, population diversity, or problem search landscape structure.

Thus, selecting operators solely on an algorithmic schedule (e.g., fixed mutation rate, fixed neighborhood size, or fixed velocity update rule) is suboptimal in energy-limited environments. We here propose EI/J to fill this gap by providing a principled ratio
between expected benefit and expected energy expenditure.

\subsection{Definition}

Let $o$ denote a metaheuristic operator (e.g., bit-flip mutation, uniform crossover, multi-bit perturbation, or a PSO velocity update). When applied at iteration $t$, the operator produces a fitness change
\begin{equation}
    \Delta f_t(o) = f(x^{(t+1)}) - f(x^{(t)}),
\end{equation}
and consumes a given amount of energy
\begin{equation}
    E_t(o) \in \mathbb{R}_{>0} \quad \text{(Joules)}.
\end{equation}

Given empirical observations from operator profiling and dynamic execution (similar to those in~\cite{Abdelhafez2019} and~\cite{Diaz2022}), we can
model the expected improvement and expected energy cost as
\begin{align}
    \mu_{\Delta f}(o) &= \mathbb{E}[\Delta f_t(o)],\\
    \mu_{E}(o)        &= \mathbb{E}[E_t(o)].
\end{align}

The \emph{Expected Improvement per Joule} can then be defined as
\begin{equation}
    \mathrm{EI/J}(o) \;=\; 
    \frac{\mu_{\Delta f}(o)}{\mu_{E}(o)}.
    \label{eq:basic_eij}
\end{equation}

Operators with higher EI/J are preferred, as they yield more expected fitness improvement per unit of energy spent. This basic estimator is later replaced by the robust EI/J estimator of Eq.~\eqref{eq:EI/J_robust} for actual operator selection.

\subsection{Estimation of Expected Improvement}
As a consequence of the previous definition, we need to measure fitness improvements (well-known) and energy consumption (not standard in research). In this second case, we would need to measure empirically and theoretically, and analyse their interplay to ground the use of this concept. One could argue that fitness improvement can also be measured theoretically and empirically, but we leave this out of our focus since there is a huge literature on it~\cite{Li2014,DURGUT2021773,Fialho2010}.

As to the expected improvement term $\mu_{\Delta f}(o)$, it can be estimated online during the algorithm's run from the history of observed fitness gains obtained whenever operator $o$ is applied.
If $\Delta f_1(o), \ldots, \Delta f_k(o)$ are the recorded improvements the:
\begin{equation}
    \mu_{\Delta f}(o)
    \approx
    \frac{1}{k} \sum_{i=1}^k \Delta f_i(o),
\end{equation}
which can be updated incrementally in $\mathcal{O}(1)$ time.


\subsection{Estimation of Expected Energy Cost}

The expected energy consumption $\mu_{E}(o)$ is here obtained from initial profiling and during execution:
\begin{equation}
    \mu_{E}(o)
    \approx
    \frac{1}{k} \sum_{i=1}^k E_i(o),
\end{equation}
where $E_i(o)$ are measurements recorded by hardware sensors accessed, e.g., through RAPL. This provides a precise, empirical estimate of the Joules needed to invoke $o$. For the present work, we safely ignore potential noise in the reading due to the nature of our algorithms, which operate on large time and energy scales.


\subsection{Thompson Sampling-based EI/J}
\label{sec:thompson}

The EI/J estimator based solely on raw sample averages may overemphasize operators that occasionally yield large improvements while exhibiting poor or highly unstable performance in most trials.
Such behaviour leads to unreliable operator ranking and suboptimal energy allocation.
To mitigate this issue, we adopt a Thompson sampling framework for energy-aware operator selection that explicitly models the uncertainty in operator performance and naturally balances exploration and exploitation.

For each operator $o$, the fitness improvement $\Delta f$ is modelled by a normal distribution, while the energy consumption $E$ is modelled by a log-normal distribution in order to account for its non-negativity and skewed empirical characteristics.
Let $\mu_{\Delta f}(o)$ and $\sigma_{\Delta f}(o)$ denote the exponentially weighted moving average (EWMA) mean and standard deviation of the observed fitness improvements, and let $\mu_{E}(o)$ and $\sigma_{E}(o)$ be the corresponding statistics for the logarithm of energy consumption.
These quantities are updated at iteration $t$ with a smoothing factor $\alpha\in(0,1)$ as
\begin{align}
\mu_{\Delta f,t}(o) &\leftarrow
    \alpha\,\mu_{\Delta f,t-1}(o)
    + (1-\alpha)\,\Delta f_t(o), \label{eq:mu_f_update}\\
\sigma_{\Delta f,t}^2(o) &\leftarrow
    \alpha\,\sigma_{\Delta f,t-1}^2(o)
    + (1-\alpha)\bigl(\Delta f_t(o)-\mu_{\Delta f,t}(o)\bigr)^2, \label{eq:sigma_f_update}\\
\mu_{E,t}(o) &\leftarrow
    \alpha\,\mu_{E,t-1}(o)
    + (1-\alpha)\ln(E_t(o)), \label{eq:mu_E_update}\\
\sigma_{E,t}^2(o) &\leftarrow
    \alpha\,\sigma_{E,t-1}^2(o)
    + (1-\alpha)\bigl(\ln(E_t(o))-\mu_{E,t}(o)\bigr)^2.\label{eq:sigma_E_update}
\end{align}
Based on these estimates, Thompson sampling draws a stochastic realization of the improvement and energy of each operator as
\begin{align}
\tilde{\Delta f}(o) &\sim \mathcal{N}\!\bigl(\mu_{\Delta f}(o), \sigma_{\Delta f}^2(o)\bigr),\label{eq:sample_f}\\
\tilde{E}(o) &\sim \mathcal{LN}\!\bigl(\mu_{E}(o), \sigma_{E}^2(o)\bigr),\label{eq:sample_E}
\end{align}
where $\mathcal{N}(\cdot)$ and $\mathcal{LN}(\cdot)$ denote the normal and log-normal distributions, respectively.
The sampled EI/J is then computed as
\begin{equation}
\mathrm{EI/J}_{\mathrm{robust}}(o) = \frac{\tilde{\Delta f}(o)}{\tilde{E}(o)}.
\label{eq:EI/J_robust}
\end{equation}
This stochastic formulation preserves the intuitive meaning of EI/J while reducing sensitivity to extreme, noisy, or sparsely observed operator outcomes, thereby providing a robust scalar quantity for comparing operators under uncertainty.

When two operators $o_1$ and $o_2$ are compared, the probability of selecting $o_1$ is interpreted as
\begin{equation}
\mathbb{P}\!\left(o_1\right)
=
\mathbb{E}_{R}
\left[
\Phi\!\left(
\frac{
\mu_{\Delta f}(o_1)-R\,\mu_{\Delta f}(o_2)
}{
\sqrt{\sigma_{\Delta f}^2(o_1)+R^2\sigma_{\Delta f}^2(o_2)}
}
\right)
\right],
\label{eq:probability}
\end{equation}
where $R=\tilde{E}(o_1)/\tilde{E}(o_2)$, and $\Phi(\cdot)$ denotes the cumulative distribution function of the standard normal.
This indicates that the scheduler selects operators based on their probability of outperforming competitors in terms of energy-normalized improvement.

The smoothing factor $\alpha$ controls the temporal adaptivity of the EWMA estimators. Smaller values of $\alpha$ (e.g., $0.1$--$0.3$) yield rapidly reactive statistics that emphasize recent observations, whereas larger values (e.g., $0.7$--$0.95$) produce more stable estimates suitable for long runs. In our experiments, values in the range $0.5$--$0.9$ were effective across GA, PSO, and ILS. In particular, a larger $\alpha$ is preferable for algorithms that apply operators frequently within each iteration, such as GA.

\subsection{Energy-aware Prioritisation under Limited Budget}
\label{sec:energy_aware_prior}

Since metaheuristics operate under a maximum available energy budget, we incorporate a multiplicative penalty that down-weights operators with high expected energy when the remaining energy $B_t$ becomes small.
For each operator $o$, the expected energy is given by the mean of the log-normal distribution, $\mathbb{E}[E(o)] = e^{\mu_E(o) + \sigma_E^2(o)/2}$.
The budget-aware penalty is then defined as
\begin{equation}
\pi(o,B_t) = \frac{B_t}{B_t + e^{\mu_E(o) + \sigma_E^2(o)/2}}.
\label{eq:budget}
\end{equation}
The final priority assigned to operator $o$ is the product
$\mathrm{EI/J}_{\mathrm{robust}}(o)\cdot\pi(o,B_t)$, ensuring that the search remains feasible while favouring operators that offer greater EI/J at the current stage.

\subsection{Operator selection rule}
\label{sec:operator_sel_rule}

At each iteration $t$, the algorithm selects the operator achieving
the largest priority value:
\begin{align}
\Phi_t(o)&=\mathrm{EI/J}_{\mathrm{robust}}(o)\cdot\pi(o,B_t),
\label{eq:priority}\\
o_t^\ast
 &=
 \arg\max_{o\in\mathcal{O}} 
 \Phi_t(o).
\label{eq:selection}
\end{align}



\section{Energy-Aware Metaheuristic Design}
This section presents the design of the proposed energy-aware metaheuristic framework, including the EI/J-based operator scheduler and its integration into GA, PSO, and ILS.

\subsection{Energy-Aware EI/J Scheduler}
\label{sec:scheduler}
The energy-aware operator scheduler (Algorithm~\ref{alg:eos}) is a central component of our framework and is used identically by GA, PSO, and ILS. At each iteration, it updates the online statistics of every operator via an exponential moving average with a smoothing factor $\alpha$, draws a stochastic EI/J realisation for each operator using Thompson sampling, and then applies a budget-aware penalty to obtain the final priority score used for selection.

Because the EI/J ratio implicitly penalizes operators with higher operation counts, the scheduler naturally favours operators whose theoretical complexity (and thus energy footprint) is lower unless the observed improvements justify their overhead.

\subsection{Energy-aware GA, PSO, and ILS}
\label{sec:energy_aware_meta}

The following pseudocodes illustrate the structure of the three metaheuristics considered in this work. Each algorithm is decomposed into two components: (i) a stable core procedure representing the traditional behaviour of the method, and (ii) a dynamic operator selection layer driven by EI/J. At each iteration, the scheduler computes the EI/J-based priority value for each operator and selects the one with the highest score. The chosen operator is then executed, and its observed improvement $\Delta f$ and the energy consumed are recorded and used to update the operator statistics. This mechanism integrates seamlessly into steady-state GA (ssGA), PSO, and ILS without modifying their canonical structure.

\begin{algorithm}[!h]
\caption{Energy-aware Operator Scheduler}
\label{alg:eos}
\small{
\begin{algorithmic}[1]

\If{$\exists o \in \mathcal{O} \text{ unselected}$}
    \State \Return a random unselected operator
\EndIf

\ForAll{$o \in \mathcal{O}$}
    \State Sample $\tilde{\Delta f}_t(o)$ and $\tilde{E}_t(o)$ as Eqs.~\eqref{eq:sample_f} and \eqref{eq:sample_E}
   \State $\mathrm{EI/J}_{\mathrm{robust}}(o) \gets \tilde{\Delta f}_t(o) / \tilde{E}_t(o)$
   \State $\pi(o,B_t) \gets B_t/\left(B_t + e^{\mu_{E}(o) + \sigma_{E}^2(o)/2}\right)$
    \State $\Phi_t(o) \gets \mathrm{EI/J}_{\mathrm{robust}}(o)\cdot\pi(o,B_t)$
\EndFor

\State \Return $o^*_t\gets \arg\max_{o \in \mathcal{O}} \Phi_t(o)$
\end{algorithmic}
}
\end{algorithm}

The ssGA shown in Algorithm~\ref{alg:ssga} maintains a fixed-size population and generates $k=1$ or $5$ offspring at a time. The scheduler selects either the low-energy \emph{Replace-1} operator or the medium-energy \emph{Replace-5} operator depending on their current EI/J values. Each operator generates $k\in\{1, 5\}$ offspring, which are merged with the current population, and the next population is obtained by truncation selection, i.e., retaining the best $n$ individuals from the union of the current population and offspring.
\begin{algorithm}[!h]
\caption{Energy-aware Steady-state GA}
\label{alg:ssga}
\small{
\begin{algorithmic}[1]

\State Initialise population $P$ of size $n$ and statistics $\mu_{\Delta f}$, $\mu_E$, $\sigma_{\Delta f}$, $\sigma_E$
\State $B_0 \gets B_{\max}$

\While{$B_t > 0$}

    \State $o \gets \textsc{SelectOperator}(\mathcal{O}, B_t)$
    \State Start energy counter
    \State $P_{\mathrm{new}} \gets \emptyset,\:\Delta f \gets 0$
    \For{$i=1$ to $k$}  \Comment{$k=1$ or $5$ decided by $o$}
        \State Select parents from $P$
        \State $x_{\mathrm{new}} \gets$ apply genetic operator to parents
        \State $\Delta f \gets \Delta f + \left(f(x_{\mathrm{new}}) - f(x_{\mathrm{ref}})\right)$\Comment{$x_{\mathrm{ref}}$ is a better parent}
        \State Insert $x_{\mathrm{new}}$ into $P_{\mathrm{new}}$
    \EndFor
    \State $P\gets \text{Top-}n(P\cup P_{\mathrm{new}})$
    \State Stop energy counter and record total energy $E$
    \State Update operator statistics using Eqs.~\eqref{eq:mu_f_update}--\eqref{eq:sigma_E_update}
   \State $B_t \gets B_t - E$

\EndWhile

\end{algorithmic}
}
\end{algorithm}


In PSO shown in Algorithm~\ref{alg:pso}, the update is identical to the canonical formulation except for the choice between two velocity-update operators. The \emph{Full} operator corresponds to the standard PSO update, while the \emph{Light} operator omits the global-best component, i.e., only the inertia and personal-best terms are used.
\begin{algorithm}[ht]
\caption{Energy-aware PSO}
\label{alg:pso}
\small{
\begin{algorithmic}[1]

\State Initialise swarm of $n$ particles and statistics $\mu_{\Delta f}$, $\mu_E$, $\sigma_{\Delta f}$, $\sigma_E$
\State $B_0 \gets B_{\max}$

\While{$B_t > 0$}

    \State $o \gets \textsc{SelectOperator}(\mathcal{O}, B_t)$
    \State Start energy counter
    \State $\Delta f \gets 0$

    \ForAll{particle $x_p$}

        \State $x_{\mathrm{new}} \gets$ apply operator $o$ to $x_p$
        \State $\Delta f \gets \Delta f + \left(f(x_{\mathrm{new}}) - f(x_p)\right)$

        \State Update personal best if needed
        \State $x_p \gets x_{\mathrm{new}}$

    \EndFor
    \State Update global best if needed
    \State Stop energy counter and record total energy $E$
    \State Update operator statistics using Eqs.~\eqref{eq:mu_f_update}--\eqref{eq:sigma_E_update}

\EndWhile

\end{algorithmic}
}
\end{algorithm}


ILS operates by iterating over perturbation and local improvement phases as presented in Algorithm~\ref{alg:ils}. 
The scheduler selects between two local search operators: a one-bit flip (\emph{ILS-1}) and a five-bit flip (\emph{ILS-5}). The chosen operator performs an initial perturbation followed by a fixed number of local improvement steps.
\begin{algorithm}[ht]
\caption{Energy-aware ILS}
\label{alg:ils}
\small{
\begin{algorithmic}[1]

\State Initialise solution $x (=x_{\mathrm{best}})$ and statistics $\mu_{\Delta f}$, $\mu_E$, $\sigma_{\Delta f}$, $\sigma_E$
\State $B_0 \gets B_{\max}$

\While{$B_t > 0$}

    \State $o \gets \textsc{SelectOperator}(\mathcal{O}, B_t)$
    \State Start energy counter
    \State $x \gets$ apply operator $o$ to $x_{\mathrm{best}}$ 
    \Statex\Comment{first perturbation (Flip 1 or 5 bits according to $o$)}
    \For{$i=1$ to $I_\mathrm{max}$}  \Comment{local improvement steps}
        \State $x' \gets$ apply operator $o$ to $x$ \Comment{Flip 1 or 5 bits according to $o$}
        \State Replace $x$ with $x'$ if $f(x')>f(x)$
    \EndFor
      \State Stop energy counter and record total energy $E$
    \State $\Delta f \gets f(x) - f(x_\mathrm{best})$
    \State Update operator statistics using Eqs.~\eqref{eq:mu_f_update}--\eqref{eq:sigma_E_update}

    \State Replace $x_\mathrm{best}$ with $x$ if $f(x)>f(x_\mathrm{best})$

    \State $B_{t+1} \gets B_t - E$

\EndWhile

\end{algorithmic}
}
\end{algorithm}

\subsection{Computational Overhead and Stability}
The proposed energy-aware framework introduces only negligible computational overhead. Measuring energy consumption requires only constant-time operations per iteration and does not change the asymptotic computational complexity of the underlying metaheuristics. The operator scheduler itself performs a small number of arithmetic operations and random number generation over a very small set of operators, resulting in $O(1)$ overhead per iteration, which is negligible in practice.

Importantly, the scheduler does not modify the internal search mechanisms of ssGA, PSO, or ILS.  It only determines which existing operator is applied at each iteration, while all other procedures, including variation, evaluation, and acceptance, remain unchanged. Consequently, the proposed framework preserves the original convergence behaviour and stability properties of the underlying algorithms and does not introduce additional sources of instability.

\section{Experimental Setup}
To evaluate the performance of the proposed energy-aware metaheuristics, we conduct comprehensive experiments using three well-known combinatorial optimisation problems. This section details the benchmark problems, energy measurement protocol, algorithm configurations, and evaluation metrics used in our experiments.

\subsection{Benchmarking Problems}

We evaluate the energy-aware metaheuristics on three combinatorial optimisation problems: the 0--1 Knapsack, NK-landscapes, and Error-Correcting Codes. These problems provide different forms of epistasis, ruggedness, and constraint structure, enabling a broad assessment of operator efficiency and EI/J scheduling.

\vspace{-0.5\baselineskip}\paragraph{0--1 Knapsack (KP)}
KP consists of selecting a subset of $n$ items with weights $w_i$ and profits $p_i$ to maximise total profit under a capacity constraint $C$. A binary vector $x \in \{0,1\}^n$ indicates whether item $i$ is selected. 
Constraint handling is performed using a penalty function
\[
f(x) = \sum_{i=1}^{n} p_i x_i - K\rho\max\left(0, \sum_{i=1}^{n} w_i x_i - C\right),
\]
where $K$ is a penalty coefficient and $\rho$ balances profits and weights.
We set $K=n$ and $\rho=\max_{i\in \{1,\dots,n\}}\left(p_i/w_i\right)$.
Experiments use the Pisinger benchmark instance~\cite{PISINGER20052271} with $n=100$, where $p_i$ and $w_i$ are independently sampled from $[1,1000]$ and $C$ is set to the maximum item weight.

\vspace{-0.5\baselineskip}\paragraph{NK-Landscapes (NK)}
NK models tunable epistasis among $n$ binary variables, where $K$ controls the number of interacting variables and thus landscape ruggedness~\cite{Kauffman1993NK}.
Fitness is defined as
\[
f(x) = \frac{1}{n} \sum_{i=1}^{n} f_i\!\left(x_i, x_{i_1}, \dots, x_{i_K}\right),
\]
with epistatic neighbourhoods $(i_1,\dots,i_K)$ of variable $i$. We use a random instance with $n=100$ and $K=6$, where each $f_i$ is uniformly sampled from $[0, 1)$, following~\cite{Pelikan2008NK}.

\vspace{-0.5\baselineskip}\paragraph{Error-Correcting Codes (ECC)}
ECC construction aims to maximise the minimum Hamming distance among $M$ codewords of length $n$~\cite{gamal1987using}. A solution $X \in \{0,1\}^{M \times n}$ is evaluated as
\[
f(X)=\frac{1}{\sum_{i=1}^{M}\sum_{j=1,j\neq i}^{M} 1/d_H(X_i,X_j)^2},
\]
where $d_H$ denotes the Hamming distance and $X_i$ is the $i$-th codeword. We consider codes with $n=12$ and $M=24$, following~\cite{ALBA2004611}.

\vspace{-0.5\baselineskip}\paragraph{Computational complexity}

Evaluation cost differs across benchmarks: KP requires $O(n)$, NK requires $O(nK)$, and ECC involves pairwise distance computations with $O(Mn^2)$.
This hierarchy provides a controlled setting to study how operator-level energy efficiency interacts with problem-dependent evaluation costs.

\subsection{Energy Measurement Protocol}
All experiments were conducted on a machine running Ubuntu~22.04 with an Intel(R) Xeon(R) CPU E5-1650 v2 operating at \SI{3.50}{\giga\hertz} and \SI{16}{\giga\byte} of DRAM.
All algorithms were implemented in Python~3.10.12 using the MEALPY library (v3.0.3)~\cite{van2023mealpy}.

Energy consumption was measured at runtime using the pyRAPL package~\cite{pyrapl_docs}, which accesses hardware RAPL counters.
For each operator execution, CPU and DRAM energy were recorded and summed as total energy consumption.
These measurements are obtained from hardware counters, rather than estimated from execution time or instruction counts.

This protocol enables operator-level energy profiling based on actual measurements, forming the foundation of the proposed EI/J-based scheduling mechanism.

\begin{figure*}[!tb]
\centering
\begin{minipage}[t]{0.32\textwidth}
\centering
   \includegraphics[width=0.8\linewidth]{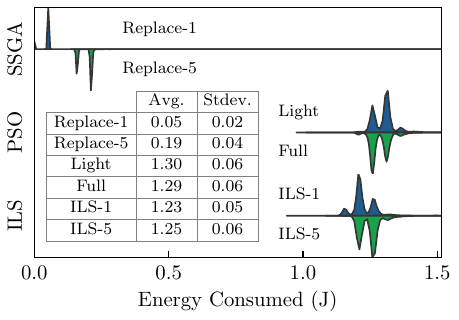}
   \subcaption{KP}
   \label{fig:KP_energy_dist}
\end{minipage}
\begin{minipage}[t]{0.32\textwidth}
\centering
   \includegraphics[width=0.8\linewidth]{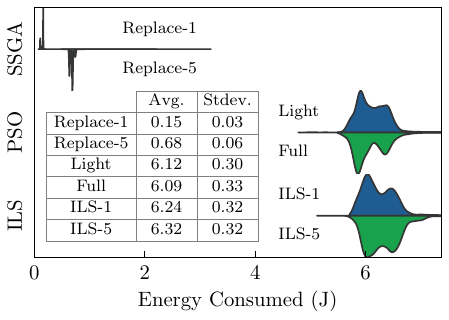}
   \subcaption{NK}
   \label{fig:NK_energy_dist}
\end{minipage}
\begin{minipage}[t]{0.32\textwidth}
\centering
   \includegraphics[width=0.8\linewidth]{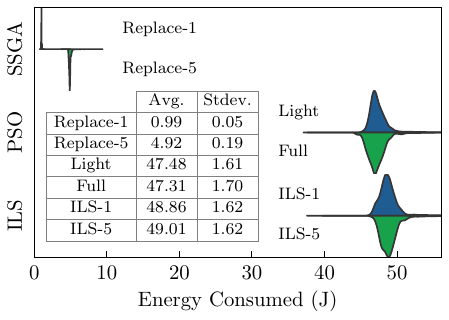}
   \subcaption{ECC}
   \label{fig:ECC_energy_dist}
\end{minipage}
\caption{Violin plot of the energy distribution per operator}
\label{fig:energy_dist}
\end{figure*}
\subsection{Parameter Settings}
\begin{table}[tb]
\caption{Algorithm parameter settings}
\label{tab:param}
\footnotesize
\centering
\setlength{\tabcolsep}{4pt}
\renewcommand{\arraystretch}{0.95}
\begin{tabular}{l@{\hspace{1.5em}}l r@{\hspace{1.5em}} l r}
\toprule
 & Parameter & Value & Parameter & Value \\
\midrule
\multirow{1}{*}{Common}
 & $\alpha$ & 0.9 &  &  \\
\midrule
\multirow{3}{*}{SSGA}
 & Population size & 100
 & Selection & Tournament \\
 & Crossover & One point
 & Crossover rate & 0.8 \\
 & Mutation & Bit flip
 & Mutation rate & $1/D$ \\
\midrule
\multirow{2}{*}{PSO}
 & Population size & 100
 & $w$ & 0.729 \\
 & $c_1$ & 2.05
 & $c_2$ & 2.05 \\
\midrule
\multirow{1}{*}{ILS}
 & Max. iter. per LS & 100
 & Acceptance & First-impr. \\
\bottomrule
\end{tabular}
\end{table}
Table~\ref{tab:param} summarises the algorithm parameters; a common smoothing factor $\alpha=0.9$ is used for the EI/J estimator, selected based on preliminary experiments that favoured more stable estimates.

SSGA uses a population size of 100 with one-point crossover (rate 0.8), bit-flip mutation (rate $1/D$), and tournament selection. 
PSO also uses a swarm size of 100, an inertia weight $w=0.729$, and cognitive and social coefficients $c_1=c_2=2.05$. 
ILS applies a first-improvement acceptance strategy with a maximum of 100 iterations.
All algorithms are evaluated over 100 independent trials. The maximum energy budget $B_{\max}$ is set to \SI{1.0}{\kilo\joule} for KP and NK, and to \SI{10.0}{\kilo\joule} for ECC.

\subsection{Initial Operator Profiling}
\label{sec:initial_profiling}

As a first step toward energy-aware optimisation, we establish baseline operator-level energy profiles for each algorithm.
Each metaheuristic is executed using a single operator variant, and the energy consumed per operator execution is recorded.
Figure~\ref{fig:energy_dist} shows the resulting distributions as violin plots. The horizontal axis denotes energy, and the vertical axis indicates each method.

In SSGA, the two operators exhibit a clear separation: Replace-5 consistently consumes more energy than Replace-1, reflecting the increased number of offspring evaluations per iteration.
In contrast, for PSO and ILS, the energy distributions of operator variants largely overlap, indicating that differences in velocity-update rules or perturbation sizes have a limited impact on the per-iteration energy cost.

Across benchmarks, absolute energy consumption increases from KP to NK and ECC, consistent with their increasing evaluation complexity.
For KP and NK, the distributions exhibit a bimodal structure with two distinct peaks. 
Based on a component-wise analysis, this pattern cannot be attributed to operator logic or fitness evaluation, suggesting the influence of system-level execution factors.
In contrast, ECC shows a unimodal distribution, indicating a more stable per-operator cost.

These results demonstrate that operators are not energy equivalent and that lightweight operator-level energy profiling already provides actionable information for guiding energy-aware selection during the search.

\subsection{Evaluation Metrics and Statistical Tests}
All methods are evaluated under fixed energy budgets, and performance is reported as solution quality per unit energy rather than per iteration. The primary outcomes are the final best-so-far fitness at budget exhaustion and the fitness--energy trajectories observed during the search.

For trajectory-level analysis, the average best-so-far fitness if fitted using a normalized saturating exponential model, $f(E) = f_\infty \bigl(1 - A e^{-k B / B_{\max}} \bigr)$. Here, $f_\infty$ represents the expected asymptotic solution quality, while $k$ captures the normalized convergence speed with respect to the energy budget, enabling comparisons across problems with different budget scales.

To assess whether the proposed energy-aware variants preserve solution quality, each EI/J-method is compared against its corresponding single-operator baselines using the Mann--Whitney U test.
In addition, operator selection ratios are reported to characterize behavioural differences induced by EI/J scheduling.

\section{Results}
This section reports the experimental results of the proposed EI/J-based metaheuristics, focusing on online estimation behaviour, operator selection, and solution quality under fixed energy budgets.

\subsection{EI/J Model Convergence and Tracking}

\begin{figure*}[tb]
\begin{minipage}[t]{0.32\textwidth}
   \includegraphics[width=\linewidth]{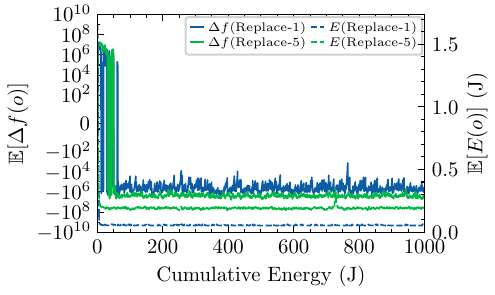}
   \subcaption{SSGA}
   \label{fig:SSGA_KP_op_energy}
\end{minipage}
\begin{minipage}[t]{0.32\textwidth}
   \includegraphics[width=\linewidth]{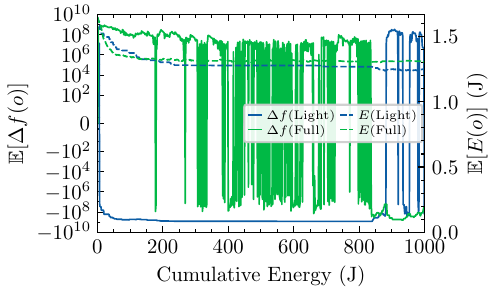}
   \subcaption{PSO}
   \label{fig:PSO_KP_op_energy}
\end{minipage}
\begin{minipage}[t]{0.32\textwidth}
   \includegraphics[width=\linewidth]{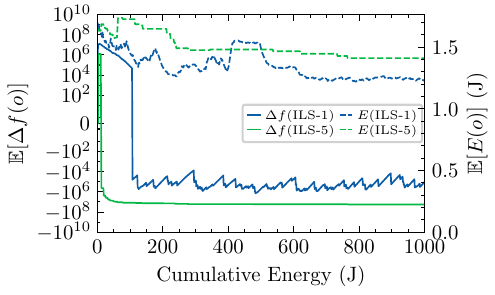}
   \subcaption{ILS}
   \label{fig:ILS_KP_op_energy}
\end{minipage}
\caption{Transitions of expected improvement and expected energy (KP)}
\label{fig:KP_op_energy}
\begin{minipage}[t]{0.32\textwidth}
   \includegraphics[width=\linewidth]{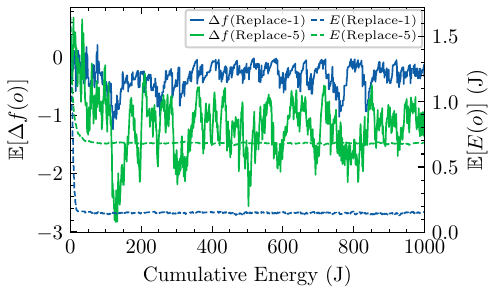}
   \subcaption{SSGA}
   \label{fig:SSGA_NK_op_energy}
\end{minipage}
\begin{minipage}[t]{0.32\textwidth}
   \includegraphics[width=\linewidth]{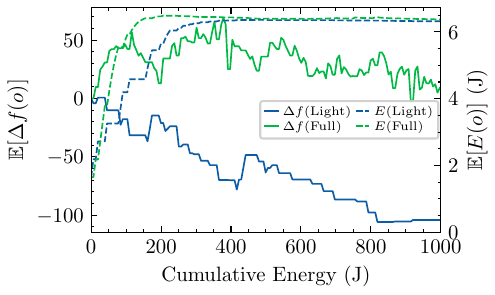}
   \subcaption{PSO}
   \label{fig:PSO_NK_op_energy}
\end{minipage}
\begin{minipage}[t]{0.32\textwidth}
   \includegraphics[width=\linewidth]{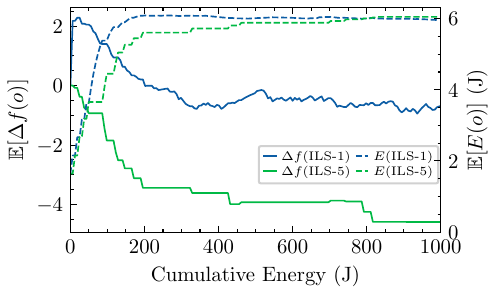}
   \subcaption{ILS}
   \label{fig:ILS_NK_op_energy}
\end{minipage}
\caption{Transitions of expected improvement and expected energy (NK)}
\label{fig:NK_op_energy}
\begin{minipage}[t]{0.32\textwidth}
   \includegraphics[width=\linewidth]{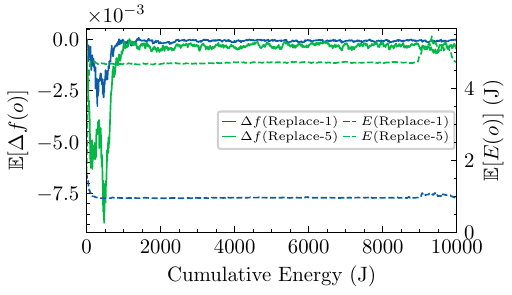}
   \subcaption{SSGA}
   \label{fig:SSGA_ECC_op_energy}
\end{minipage}
\begin{minipage}[t]{0.32\textwidth}
   \includegraphics[width=\linewidth]{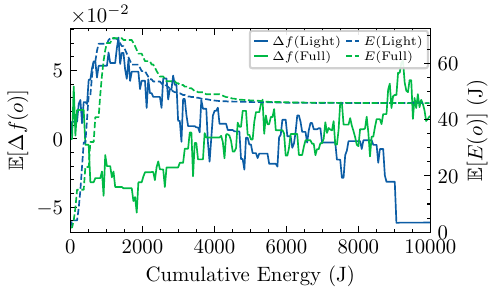}
   \subcaption{PSO}
   \label{fig:PSO_ECC_op_energy}
\end{minipage}
\begin{minipage}[t]{0.32\textwidth}
   \includegraphics[width=\linewidth]{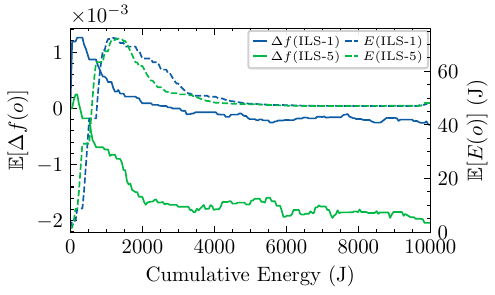}
   \subcaption{ILS}
   \label{fig:ILS_ECC_op_energy}
\end{minipage}
\caption{Transitions of expected improvement and expected energy (ECC)}
\label{fig:ECC_op_energy}
\end{figure*}

\begin{figure}[tb]
\includegraphics[width=0.32\textwidth]{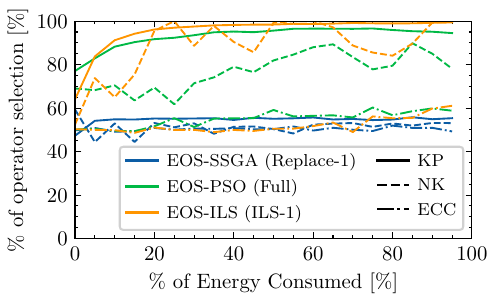}
\caption{Operator selection ratios across energy budget (\%)}
\label{fig:sel_ratio}
\end{figure}
\begin{figure*}[!tb]
\begin{minipage}[t]{0.32\textwidth}
   \includegraphics[width=\linewidth]{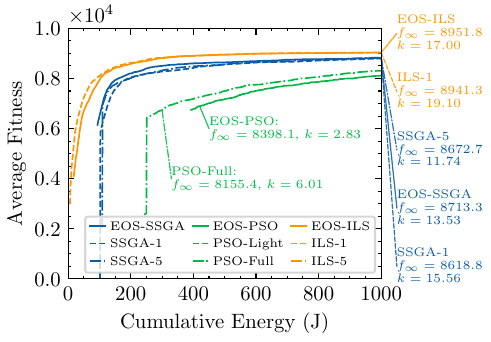}
   \subcaption{KP}
   \label{fig:KP_fitness}
\end{minipage}
\begin{minipage}[t]{0.32\textwidth}
   \includegraphics[width=\linewidth]{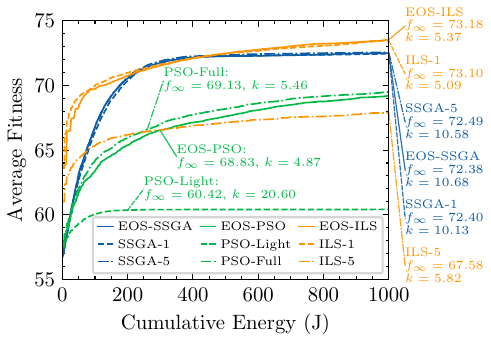}
   \subcaption{NK}
   \label{fig:NK_fitness}
\end{minipage}
\begin{minipage}[t]{0.32\textwidth}
   \includegraphics[width=\linewidth]{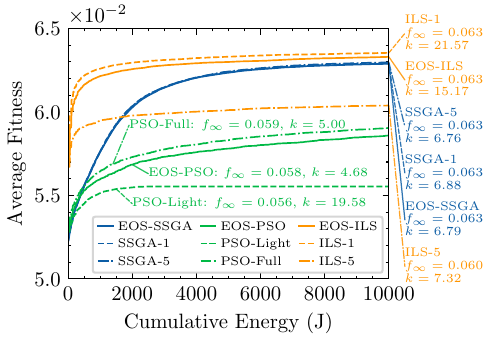}
   \subcaption{ECC}
   \label{fig:ECC_fitness}
\end{minipage}
\caption{Trajectories of average fitness via cumulative energy consumption (\si{\joule})}
\label{fig:fitness_transition}
\end{figure*}
Figures~\ref{fig:KP_op_energy}--\ref{fig:ECC_op_energy} show the evolution of the estimated expected improvement $\mathbb{E}[\Delta f(o)]$ and expected energy consumption $\mathbb{E}[E(o)]$ plotted against cumulative energy for representative median trials.

Across all algorithms and problems, the estimated trajectories reveal a consistent coupling between numerical improvement and energy consumption at the operator level.
Operators that incur higher expected energy do not necessarily yield larger expected improvements, and meaningful differences in $\mathbb{E}[\Delta f(o)]$ can be observed even when $\mathbb{E}[E(o)]$ remains comparable.

The online estimators remain stable as energy accumulates.
In particular, $\mathbb{E}[E(o)]$ closely matches the empirical operator profiles reported in Section~\ref{sec:initial_profiling}, indicating that the learned statistics are grounded in actual measurements rather than short-term noise.
As a result, both $\mathbb{E}[\Delta f(o)]$ and $\mathbb{E}[E(o)]$ converge toward smooth, informative signals suitable for energy-aware decision making.

Differences in the scale and smoothness of $\mathbb{E}[\Delta f(o)]$ reflect how each metaheuristic accumulates improvements—over offspring in SSGA, over the swarm in PSO, or over fixed local-search steps in ILS.
Despite these structural differences, the EI/J estimation process remains effective and comparable across all metaheuristics.

\subsection{Operator Selection Dynamics}

Figure~\ref{fig:sel_ratio} reports the evolution of average operator selection ratios as a function of the consumed energy budget.
The horizontal axis represents the percentage of the energy budget used, while the vertical axis shows the probability of selecting a given operator.
For SSGA, the Replace-1 ratio is reported; for PSO and ILS, the Full and ILS-1 ratios are shown, respectively.

Across all algorithms, operator preferences emerge progressively rather than suddenly.
Even when an operator becomes dominant, its selection ratio remains around 60\% in the early stages and increases progressively, indicating online learning rather than fixed bias.

In SSGA, the selection ratio remains close to 50\% across all problems, reflecting the fact that Replace-1 and Replace-5 share the same underlying replacement mechanism and differ mainly in the number of offspring generated.
In contrast, PSO and ILS exhibit problem-dependent convergence toward a dominant operator: in KP and NK, Full and ILS-1 dominate, respectively, whereas in ECC, the selection ratios remain closer to parity.
This behaviour is explained by the large $\sigma^2_{\Delta f,t}(o)$ relative to $\mu_{\Delta f, t}(o)$, leading to comparable selection probabilities as described in Eq.~\eqref{eq:probability}. This indicates that EI/J-based scheduling naturally balances exploitation and exploration by accounting not only for expected improvement, but also for uncertainty in operator performance.

\subsection{Performance Under Fixed Energy Budgets}

Figure~\ref{fig:fitness_transition} reports fitness trajectories via cumulative energy consumption under fixed energy budgets. The horizontal axis shows cumulative energy, and the vertical axis shows fitness.
This setting allows a direct comparison of solution quality per unit energy across algorithms.
For KP, only feasible trials are shown, as some trials in two baselines (PSO-Light and ILS-5) failed to find feasible solutions.

Across all benchmarks, a consistent pattern emerges.
Static operator choices can lead to performance degradation or even failure, as observed for PSO-Light and ILS-5.
In contrast, the proposed EI/J-based methods (EOS-x) remain robust across problems, producing feasible solutions and stable improvement trajectories without prior knowledge of the most suitable operator.

When static operators perform well (e.g., PSO-Full or ILS-1), EOS-x closely matches their convergence behaviour.
Conversely, when static choices are suboptimal, EOS-x avoids catastrophic failure and maintains steady progress.
As a result, EOS-x achieves solution quality per unit energy comparable to the best-performing static configuration across benchmarks, while exhibiting lower variance across runs, indicating more stable and robust behaviour under fixed energy budgets.

This observation is supported by the fitted trajectory parameters.
Although PSO-Full and ILS-1 achieve the best combinations of asymptotic performance $f_\infty$ and convergence rate $k$, the corresponding EOS variants attain values of $f_\infty$ and $k$ that are consistently close to those of the best single-operator configurations.
This indicates that, from the perspective of the fitted convergence behaviour, consistently selecting the empirically best operator is not strictly necessary to achieve near-optimal performance under fixed energy budgets.
Instead, EOS-x reproduces most of the benefits of the best operator in terms of both final solution quality and convergence dynamics, despite occasionally selecting suboptimal operators.

These results demonstrate that energy-aware operator selection is a necessary mechanism for reliable search under fixed energy budgets.
By adapting operator usage online based on measured energy and observed improvements, EOS-x provides a problem-independent mechanism to balance numerical progress and computational cost.

\subsection{Solution Quality and Statistical Comparison}

Table~\ref{tb:final_fitness} reports the mean and standard deviation of the final fitness values obtained by each method, together with the corresponding $p$-values.
Bold values indicate the highest fitness among methods that are not statistically different at the 5\% significance level.
For KP, results for PSO-Light and ILS-5 are omitted, as these methods failed to produce feasible solutions in most trials.

For SSGA, solution quality remains statistically unchanged across EOS-SSGA, SSGA-1, and SSGA-5 for all problems, indicating that energy-aware operator selection does not degrade performance.
For PSO, EOS-PSO consistently outperforms the weaker baseline (PSO-Light) and achieves solution quality comparable to PSO-Full on KP and NK, with PSO-Full showing an advantage only on ECC.
For ILS, EOS-ILS matches the performance of ILS-1 on KP and NK while outperforming ILS-5 across all problems; only on ECC does ILS-1 yield significantly better results.

Overall, the EI/J-based method attains solution quality comparable to that of the best single-operator configurations in 7 out of 9 comparisons.
At the same time, it significantly outperforms weaker static operators (PSO-Light and ILS-5) in all 6 applicable cases.
In addition, for EOS-ILS in KP and NK, the mean fitness exceeds that of ILS-1, even though no significant difference is found.
These results demonstrate that energy-aware operator selection can preserve, and often improve, solution quality without requiring prior knowledge of the most suitable operator for a given problem.

\begin{table}[tb]
\centering
\caption{Mean and standard deviation of the final fitness}
\label{tb:final_fitness}
\scriptsize{
\begin{tabular}{l@{\hspace{1em}}r@{\hspace{2.5pt}}r@{\hspace{1em}}r@{\hspace{2.5pt}}r@{\hspace{1em}}r@{\hspace{2.5pt}} r}
\toprule
 & \multicolumn{2}{c}{KP} & \multicolumn{2}{c}{NK} & \multicolumn{2}{c}{ECC ($\times 10^{-2}$)} \\
 \cmidrule(lr){2-3}\cmidrule(lr){4-5}\cmidrule(lr){6-7}
Method & $\text{Avg.}\pm\text{Stdev.}$ & $p$-val. & $\text{Avg.}\pm\text{Stdev.}$ & $p$-val.& $\text{Avg.}\pm\text{Stdev.}$ & $p$-val.\\
\midrule
EOS-SSGA & $\mathbf{8818.55 \pm 261.56}$&--- & $\mathbf{72.49 \pm 1.36}$&--- & $\mathbf{6.29 \pm 0.03}$&--- \\
SSGA-1 & $\mathbf{8793.64 \pm 298.88}$& $0.66$ & $\mathbf{72.46 \pm 1.46}$& $0.68$ & $\mathbf{6.29 \pm 0.03}$& $0.64$ \\
SSGA-5 & $\mathbf{8834.12 \pm 284.36}$& $0.50$ & $\mathbf{72.55 \pm 1.28}$& $0.79$ & $\mathbf{6.30 \pm 0.03}$& $0.05$ \\
\midrule
EOS-PSO & $\mathbf{8120.82 \pm 667.06}$&--- & $\mathbf{69.19 \pm 1.62}$& ---& $5.86 \pm 0.06$&--- \\
PSO-Light & ---& $<0.01$ & $60.42 \pm 1.03$& $<0.01$ & $5.55 \pm 0.07$& $<0.01$ \\
PSO-Full & $\mathbf{8307.06 \pm 524.74}$& $0.07$ & $\mathbf{69.50 \pm 1.39}$& $0.11$ & $\mathbf{5.90 \pm 0.06}$& $<0.01$ \\
\midrule
EOS-ILS & $\mathbf{9035.58 \pm 125.06}$&--- & $\mathbf{73.52 \pm 1.24}$&--- & $6.33 \pm 0.06$&--- \\
ILS-1 & $\mathbf{9031.08 \pm 112.59}$& $0.91$ & $\mathbf{73.45 \pm 1.31}$& $0.53$ & $\mathbf{6.35 \pm 0.06}$& $<0.01$ \\
ILS-5 & ---& $<0.01$ & $67.89 \pm 0.94$& $<0.01$ & $6.04 \pm 0.03$& $<0.01$ \\
\bottomrule
\end{tabular}

}
\end{table}

\subsection{Energy Savings and Behavioural Differences}
\begin{table}[tb]
\centering
\caption{Energy consumption under fixed evaluation budgets}
\label{tb:final_energy}
\scriptsize{
\begin{tabular}{l@{\hspace{1em}}r@{\hspace{2.5pt}}r@{\hspace{1em}}r@{\hspace{2.5pt}}r@{\hspace{1em}}r@{\hspace{2.5pt}} r}
\toprule
 & \multicolumn{2}{c}{KP} & \multicolumn{2}{c}{NK} & \multicolumn{2}{c}{ECC} \\
 \cmidrule(lr){2-3}\cmidrule(lr){4-5}\cmidrule(lr){6-7}
 Method 
& $\text{Avg.}\pm\text{Stdev.}$ (\si{\joule}) & $p$-val. 
& $\text{Avg.}\pm\text{Stdev.}$ (\si{\joule}) & $p$-val.
& $\text{Avg.}\pm\text{Stdev.}$ (\si{\joule}) & $p$-val.\\
\midrule
EOS-SSGA 
& $688.20 \pm 55.58$ & --- 
& $\mathbf{686.57 \pm 30.17}$ & --- 
& $\mathbf{7348.27 \pm 135.92}$ & --- \\
SSGA-1 
& $798.84 \pm 30.42$ & $<0.01$ 
& $731.25 \pm 37.62$ & $<0.01$ 
& $7446.86 \pm 174.90$ & $<0.01$ \\
SSGA-5 
& $\mathbf{679.98 \pm 19.47}$ & $<0.01$ 
& $\mathbf{681.25 \pm 27.80}$ & 0.08 
& $\mathbf{7379.68 \pm 152.72}$ & 0.19 \\
\midrule
EOS-PSO 
& $\mathbf{454.99 \pm 13.59}$ & --- 
& $\mathbf{604.49 \pm 26.23}$ & --- 
& $\mathbf{7062.47 \pm 143.89}$ & --- \\
PSO-Light 
& $\mathbf{455.61 \pm 13.86}$ & 0.49 
& $\mathbf{607.44 \pm 25.85}$ & 0.36 
& $\mathbf{7080.38 \pm 134.86}$ & 0.43 \\
PSO-Full 
& $\mathbf{452.67 \pm 12.52}$ & 0.13 
& $606.17 \pm 29.10$ & 0.69 
& $\mathbf{7056.78 \pm 151.06}$ & 0.64 \\
\midrule
EOS-ILS 
& $\mathbf{425.98 \pm 12.02}$ & --- 
& $\mathbf{615.23 \pm 24.50}$ & --- 
& $\mathbf{7197.76 \pm 133.57}$ & --- \\
ILS-1 
& $\mathbf{426.45 \pm 14.77}$ & 0.41 
& $\mathbf{614.85 \pm 28.18}$ & 0.65 
& $\mathbf{7194.98 \pm 134.27}$ & 0.78 \\
ILS-5 
& $433.55 \pm 11.76$ & $<0.01$ 
& $\mathbf{621.95 \pm 27.10}$ & 0.09 
& $\mathbf{7220.81 \pm 137.15}$ & 0.24 \\
\bottomrule
\end{tabular}
}
\end{table}
Table~\ref{tb:final_energy} summarises the energy consumption (\si{\joule}) under fixed evaluation budgets, set to 35{,}000 for KP, 10{,}000 for NK, and 15{,}000 for ECC.
Bold values indicate the lowest mean energy among methods that are not statistically different at the 5\% significance level.

EOS-x achieves the lowest energy consumption or a statistically equivalent level in 8 out of 9 comparisons.
The only exception is SSGA on KP, where SSGA-5 attains slightly lower energy usage.
These results indicate that the proposed EI/J-based scheduler can effectively reduce energy consumption while preserving solution quality under a fixed number of fitness evaluations.

EOS-SSGA achieves an energy efficiency comparable to that of SSGA-5, despite selecting the more energy-intensive operator (Replace-1) in approximately half of the iterations (Figure~\ref{fig:sel_ratio}).
This suggests that the EI/J-based scheduler does not optimise operator-level energy consumption in isolation, but instead adapts operator usage to mitigate inefficient algorithm-level execution patterns.

Overall, these results suggest that the proposed EI/J-based scheduler can provide competitive energy efficiency across different problems without prior knowledge of the most suitable operator.


\section{Conclusions and Future Work}

This paper introduced a unified framework for energy-aware metaheuristics based on a robust numerical Expected Improvement per Joule (EI/J) model, enabling principled operator selection under fixed energy budgets. The approach combines operator profiling, lightweight/heavy operator variants, and a general scheduler applicable to GA, PSO, and ILS.

The energy-aware algorithms designed consistently preserved or improved solution quality while requiring less energy than their standard counterparts. EI/J stabilized early, identified the most efficient operators, and guided the search without degrading numerical performance across the three different problems.

The framework is directly portable to other operator-driven methods such as EDAs, VNS, ALNS, or multi-operator EA hybrids.

Future work may extend EI/J to larger operator portfolios and adaptive schemes such as bandits or reinforcement learning, explore energy-aware control of population and algorithmic parameters, and apply the framework to parallel or real-world large-scale settings. These directions suggest a new generation of metaheuristics that treat energy as a primary design constraint, dynamically selecting operators and search behaviours based on their improvement-per-Joule efficiency, and enabling algorithms that remain effective even under strict computational or sustainability requirements like the ones needed in mobile phones, federated learning, or in kernel techniques for generative AI training.

\begin{acks}
During the preparation of this work, the authors used Word Processing tools, Spreadsheets, and AI assistants in order to improve the readability of the text and figures. After that, the authors reviewed and edited the content as needed and then took full responsibility for the published article.
\end{acks}

\section*{Disclaimer}
This paper does not represent the opinion of any of our affiliated organisations.
\bibliographystyle{ACM-Reference-Format}
\bibliography{references}

\end{document}